\documentclass[11pt,a4paper]{article}
\usepackage[hyperref]{naaclhlt2018}
\usepackage{times}
\usepackage{latexsym}
\usepackage{graphicx} 
\usepackage{float}
\usepackage{amsmath}
\usepackage{amssymb}
\usepackage{multirow}
\usepackage{booktabs}
\usepackage{array}
\usepackage{url}
\usepackage{enumitem}

\aclfinalcopy 

\title{HFL-RC System at SemEval-2018 Task 11: Hybrid Multi-Aspects Model for Commonsense Reading Comprehension}
\author{
  Zhipeng Chen$^*$$^\dag$,Yiming Cui$^*$$^\dag$, Wentao Ma$^\dag$, Shijin Wang$^\dag$, Ting Liu$^\ddag$ \and Guoping Hu$^\dag$\\
  {$^\dag$Joint Laboratory of HIT and iFLYTEK (HFL), iFLYTEK Research, Beijing, China}\\
  {$^\ddag$Research Center for Social Computing and Information Retrieval (SCIR),}\\
  {Harbin Institute of Technology, Harbin, China}\\
  {$^\dag$\tt\{zpchen,ymcui,wtma,sjwang3,gphu\}@iflytek.com}\\
  {$^\ddag$\tt tliu@ir.hit.edu.cn}
   }
\date{}

\begin{document}
\maketitle
\begin{abstract}
This paper describes the system which got the state-of-the-art results at SemEval-2018 Task 11: Machine Comprehension using Commonsense Knowledge.
In this paper, we present a neural network called Hybrid Multi-Aspects (HMA) model, which mimic the human's intuitions on dealing with the multiple-choice reading comprehension.
In this model, we aim to produce the predictions in multiple aspects by calculating attention among the text, question and choices, and combine these results for final predictions.
Experimental results show that our HMA model could give substantial improvements over the baseline system and got the first place on the final test set leaderboard with the accuracy of 84.13\%.
\end{abstract}

\renewcommand{\thefootnote}{}
\footnotetext{\noindent$^*$Equal contribution.}
\renewcommand{\thefootnote}{\arabic{footnote}}

\section{Introduction}
Machine Reading Comprehension (MRC) has become a spotlight topic in recent natural language processing field. MRC consists of various subtasks, such as cloze-style reading comprehension \citep{hermann-etal-2015,hill-etal-2015,cui-etal-2016-consensus,cmrc2017-dataset}, span-extraction reading comprehension \citep{rajpurkar-etal-2016} and open-domain reading comprehension \citep{chen-etal-2017}, etc.
One key problem in reading comprehension is that how the machine utilizeS the commonsense knowledge for real-life reading comprehension.
In the SemEval-2018 Task 11: Machine Comprehension using Commonsense Knowledge \citep{SemEval2018Task11}, the organizers provide narrative texts about everyday activities and require the participants to build a system for answering questions based on this text.
To tackle this problem, in this paper, we present a novel model called Hybrid Multi-Aspects (HMA) model.
The main features of our model can be concluded as follows.
\begin{itemize}
	\item Our model is mainly based on the neural network approach without using any external knowledge, such as script knowledge, etc.
	\item We aim to produce the predictions in multiple aspects by calculating attention among the text, question and choices, and combine these results for final predictions.
	\item We add additional features on the embedding representations for the text, question and choices, including word matching feature and part-of-speech tags, etc.
\end{itemize}

\section{System Description}
\begin{figure*}
	\centering \includegraphics [width=0.98\textwidth]{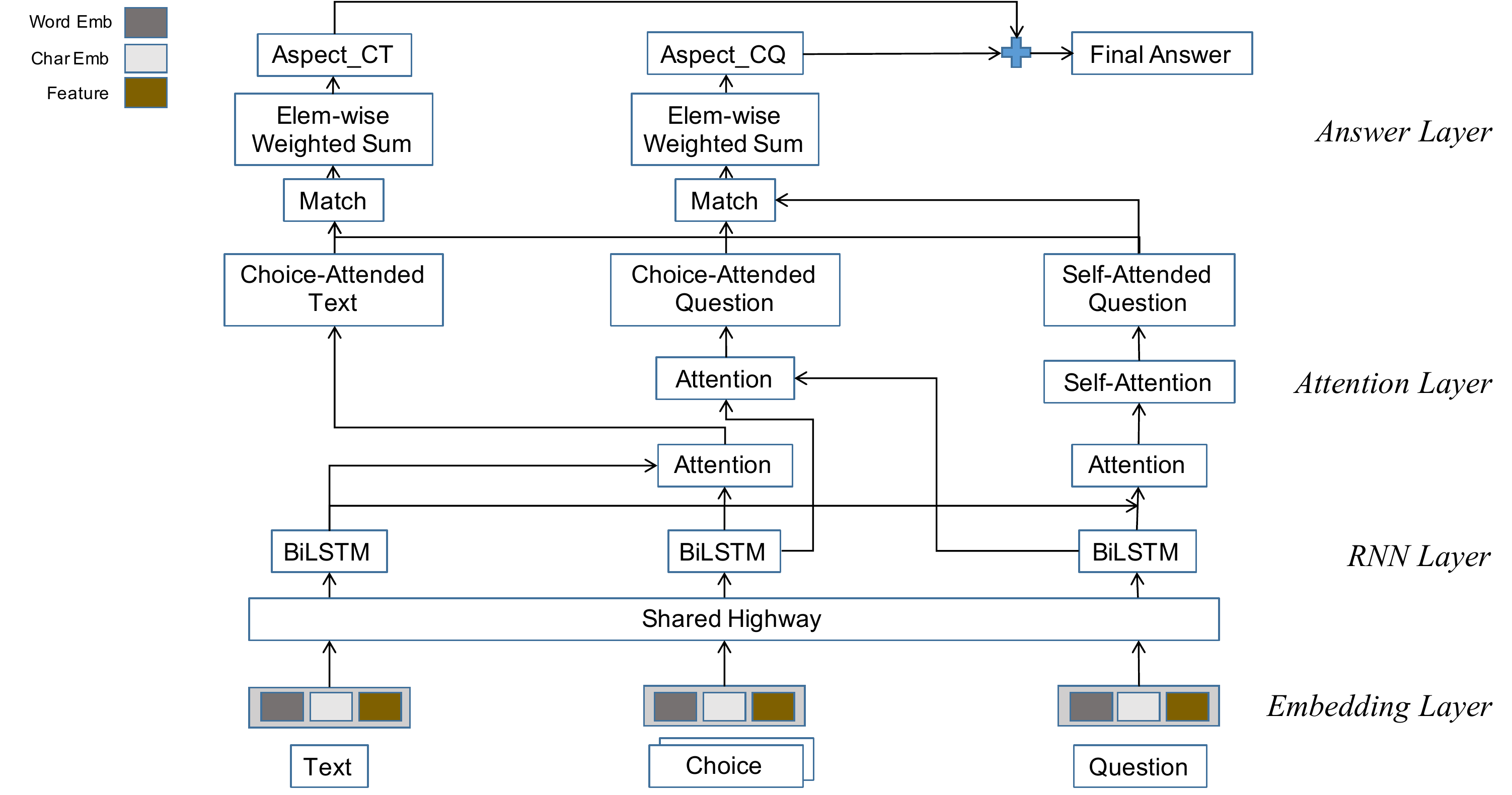}
	\caption{Neural architecture of our Hybrid Multi-Aspects (HMA) model. Note that, there are two choices processed in parallel.}\label{fig1} 
\end{figure*}

We will first give a brief introduction of the SemEval-2018 Task 11.
Then the pre-processing and the proposed Hybrid Multi-Aspects model will be illustrated afterwards. 
A quick glance of the neural architecture of the HMA model is depicted in Figure \ref{fig1}.

\subsection{Task Description}
Given a short context about the narrative texts about everyday activities and several following questions about the context, the participants are required to build a system to solve the question by choosing the correct answer from two choice choices.
The participants are encouraged to use external knowledge to improve their systems and there is no restrictions.
For more details, please infer \citet{SemEval2018Task11}.

\subsection{Pre-processing}
We describe the pre-processing procedure on the evaluation data, which can be listed as follows. 
\begin{enumerate}
 \item All punctuations are removed.
 \item All words are lower-cased.
 \item All sentences are tokenized by Natural Language Toolkit (NLTK) \citep{nltk}.
\end{enumerate}

\subsection{Embedding Layer}
In this layer, we aim to project text, question and choices into embedding representations.
The final embedding representations are composed by three components, which can be listed as follows.
\begin{enumerate}
	\item {\bf Word embedding}: We use pre-trained GloVe embedding \citep{pennington-etal-2014} for word representations, whose size is 100d ({\em d} for dimension). 
	\item {\bf Char embedding}: We use randomly initialized embedding matrix for char-level embeddings, whose size is 8d. We use 1D-convolution operation with filter length of 5 and output size of 100d. Then we apply max-over-time-pooling to obtain the final representation, whose size is 100d.
	\item {\bf Feature embedding}: 
	We also adopt several hand-crafted features for enhancing the word representations.
	In this paper, we adopt three features which can be illustrated as follows.
	\begin{enumerate}
		\item {\bf Part-of-Speech}: We use NLTK \citep{nltk} for part-of-speech tagging for each word in the text, question and choices. In this paper, we assign different trainable vectors with size of 16d for each part-of-speech tag.
		\item {\bf Word matching}: Taking the text as an example, if the text word appear in question or choice, we set this feature as 1. If not, set it as 0. In this way, we can also add this feature to the question and choice.
		\item {\bf Word fuzzy matching}: Similar to the `word matching' feature, but we loosen the matching criteria as partial matching. For example, we regard `teacher' and `teach' as fuzzy matching, because the string `teacher' is partially matched by `teach'.
	\end{enumerate}
\end{enumerate}

After obtaining three parts of the embeddings, the final representation is the concatenation of three embeddings, forming the size of 100d+100d+16d+2d=218d.

\subsection{RNN Layer}
After obtaining the embedding representations $E$, we first feed this into a shared Highway network \citep{srivastava-etal-2015} across the text, question and choices.
The output activation is chosen as $\tanh$.
\begin{equation}
	HW = \tanh(Highway(E))
\end{equation}

After highway network, we get text, question and choices' presentation $HW_{T}\in {\mathbb{R}}^{t \times e}$, $HW_{Q}\in \mathbb{R}^{q \times e}$ and $HW_{C}\in \mathbb{R}^{cn\times c \times e}$, where $t$, $q$, $c$ and $cn$ represent the text max length, question max length , choice max length and the number of choices respectively. 
Then we use Bi-directional LSTM \citep{graves-2005} and concatenate the forward and backward hidden representations to obtain the contextual representations of text $B_{T}\in {\mathbb{R}}^{t \times h}$, question $B_{Q}\in \mathbb{R}^{q \times h}$ and choices $B_{C}\in \mathbb{R}^{cn\times c \times h}$, where $h$ presents Bi-LSTM hidden size. Note that, we use separated Bi-LSTM (without sharing weights) for the text, question and choices.
\begin{gather}
	B_{T}=\text{Bi-LSTM}(HW_{T}) \\
	B_{Q}=\text{Bi-LSTM}(HW_{Q}) \\
	B_{C}=\text{Bi-LSTM}(HW_{C})
\end{gather}

\subsection{Attention Layer}
After obtaining the contextual representations of the text, question and choices, we will calculate the attentions between different combinations in order to characterize the choices in multiple aspects.
In this paper, we aim to obtain three representations
\begin{enumerate}
	\item choice-aware text representation $H_{CT}$
	\item choice-aware question representation $H_{CQ}$
	\item self-attended question representation $H_{QQ}$
\end{enumerate}

First, we will calculate choice-aware text representation $H_{CT}$ to extract the choice-relevant part from text representation. 
Following \citet{cui-acl2017-aoa}, we first calculate dot similarity between each word in the text and choice to obtain the matching matrix.
Then we apply row-wise softmax to obtain individual text-level attention with respect to each choice word, denoted as $M_{CT}\in \mathbb{R}^{c \times t}$.
\begin{gather}
	M^{'}_{CT} = B_{C} \cdot B_{T}^{T} \\
	M_{CT} = softmax(M^{'}_{CT})
\end{gather}

Then we will get choice-aware text representation $H_{CT}\in \mathbb{R}^{c \times 2h}$. Note that, we will also concatenate choice LSTM representation $B_C$ to enhance the representation power.
\begin{gather}
	H^{'}_{CT} = M_{CT} \cdot B_{T} \\
	H_{CT} = concat [H^{'}_{CT} ~;~ B_C]
\end{gather}

In the same way, we can also obtain the choice-aware question representation $H_{CQ}\in \mathbb{R}^{c \times 2h}$ and the question-aware text representation $H_{QT}\in \mathbb{R}^{q \times 2h}$.
However, in order to get further extract question information, we apply additional self-attention procedure to obtain self-attention matching matrix $M_{QQ}\in \mathbb{R}^{q \times q}$.
\begin{gather}
	M^{'}_{QQ} = H_{QT} \cdot H_{QT}^{T} \\
	M_{QQ} = softmax(M^{'}_{QQ})
\end{gather}
Then we will get self-attended question representation $H_{QQ}\in \mathbb{R}^{q \times 2h}$.
\begin{gather}
	H^{'}_{QQ} = M_{QQ} \cdot B_{Q} \\
	H_{QQ} = concat [H^{'}_{QQ} ~;~ B_q]
\end{gather}

\subsection{Answer Layer}
In this module, we will utilize the multi-aspect representations from previousl layer to get a hybrid prediction.
First, we will calculate the similarity between the choice-aware text representation $H_{CT}$ and self-attended question representation $H_{QQ}$ to obtain the deep matching matrix $D_{CT}$.
Similarly, we can also calculate the similarity between the choice-aware question representation $H_{CQ}$ and self-attended question representation $H_{QQ}$ to obtain the deep matching matrix $D_{CQ}$.
\begin{gather}
	D_{CT} = H_{CT} \cdot H^{T}_{QQ} \\
	D_{CQ} = H_{CQ} \cdot H^{T}_{QQ}
\end{gather}

Then we will apply a element-wise weight $W_{CT} \in \mathbb{R}^{c \times q}$ and get the weighted sum of $D_{CT}$ and output a scalar value. Note that, we have two choices, so the final output should be $A_{CT}\in \mathbb{R}^{1 \times 2}$.
In the same way, we can also calculate $A_{CQ}\in \mathbb{R}^{1 \times 2}$.
\begin{gather}
	A_{CT} = \sum D_{CT} \odot W_{CT} \\
	A_{CQ} = \sum D_{CQ} \odot W_{CQ}
\end{gather}

Finally, we apply softmax function to $A_{CT}$ and $A_{CQ}$ and sum the probabilities to get final predictions $A$.
\begin{gather}
	A = softmax(A_{CT}) + softmax(A_{CQ})
\end{gather}

\subsection{Training Criterion}
We use categorical cross entropy to calculate loss between the predicted answer probability $A$ and real answer.

\section{Experiments}
\subsection{Experimental Setups}
We listed the main hyper-parameters of our model in Table \ref{hyper-params}.
The word embeddings are initialized by the pre-trained GloVe word vectors (Common Crawl, 6B tokens, 100-dimension) \citep{pennington-etal-2014}. The words that do not appear in the pre-trained word vectors are set to the `unk' token and initialized accordingly.
We use Adam for weight optimizations with default parameters. 
The models are built on Keras \citep{chollet2015keras} with Theano backend \citep{theano2016}. 
We choose our model by the performance of the development set.

\begin{table}[htp]
\begin{center}
\begin{tabular}{l l c}
\toprule
{\bf Symbol} & {\bf Descriptions} & {\bf \#} \\
\midrule
{\em t} & Text max length & 300  \\ 
{\em q} & Question max length & 20  \\ 
{\em cn} & Number of choices & 2   \\ 
{\em e} & Embedding size & 218   \\ 
{\em h} & Bi-LSTM output size & 200   \\ 
{\em c} & choice max length & 10  \\
\bottomrule
\end{tabular}
\end{center}
\caption{\label{hyper-params} Hyper-parameter settings of our system.}
\end{table}

\subsection{Results}
The experimental results are shown in Table \ref{overall-results}.
We also listed some of the top-ranked systems in this evaluation\footnote{Full SemEval-2018 Task 11 leaderboard can be accessed through: \url{https://competitions.codalab.org/competitions/17184\#results}}.
As we can see that our single model could give substantial improvements over Simple RNN baseline system by 12.78\% on development set.
For further improve the performance, we carried out 7-ensemble by voting approach produced by each single system. 
The results show that the ensemble system could give further improvements where 1.98\% and 3.19\% gains on development and test set respectively.
When compared to the several top-ranked systems, our HMA model surpasses all the competitors and got on the first place in SemEval-2018 Task 11.

\begin{table}[ht]
\begin{center}
\begin{tabular}{lcc}
\toprule
\bf Model & \bf Dev & \bf Test \\ 
\midrule
Simple RNN & 71.70\% & - \\ 
HMA Model (single) & 84.48\% & 80.94\% \\ 
HMA Model (ensemble) & 86.46\% & {\bf 84.13\%} \\
\midrule
Yuanfudao (BananaTree) & - & 83.95\% \\
MITRE (guidoz) & - & 82.27\% \\
(jiangnan) & - & 80.91\% \\
Rusalka (minerva) & - & 80.48\% \\
SLQA (mingyan) & - & 79.94\% \\
\bottomrule
\end{tabular}
\end{center}
\caption{\label{overall-results} Experimental results. Top-ranked participant's systems are also included.}
\end{table}

\section{Discussion}
\begin{figure}
	\centering \includegraphics [width=0.95\linewidth]{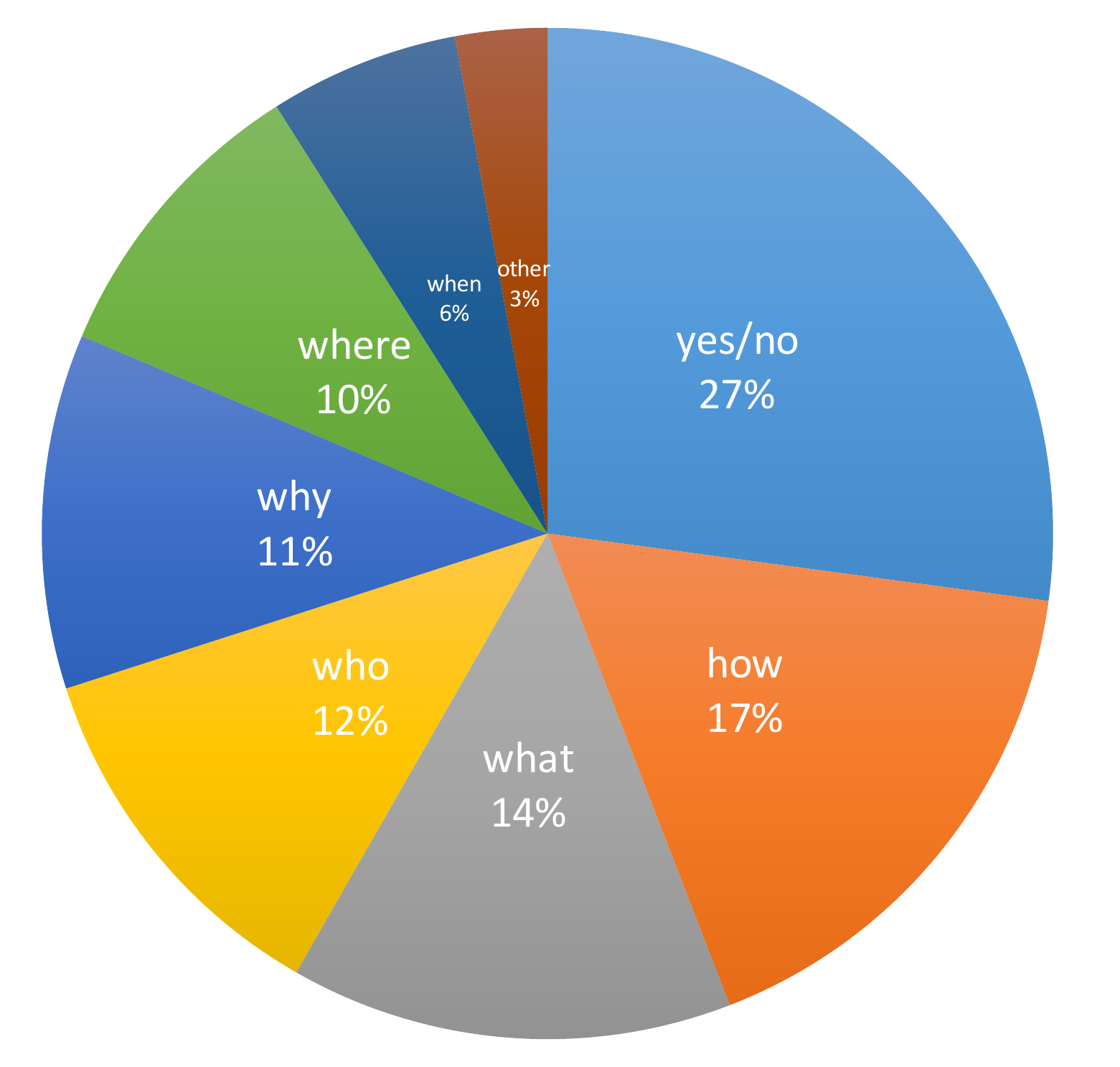}
	\caption{Proportions of different question types.}\label{fig3} 
\end{figure}

To have a better understanding of the data, we divided the data into different question types, which can be seen in Figure \ref{fig3}.
As we can see that the yes/no question takes up 27\% proportion, which is quite different from most of the other reading comprehension datasets. The yes/no questions require better handling the negation and deeper understanding in question. 

\begin{figure}[tbp] 
\centering
\includegraphics[width=0.95\linewidth]{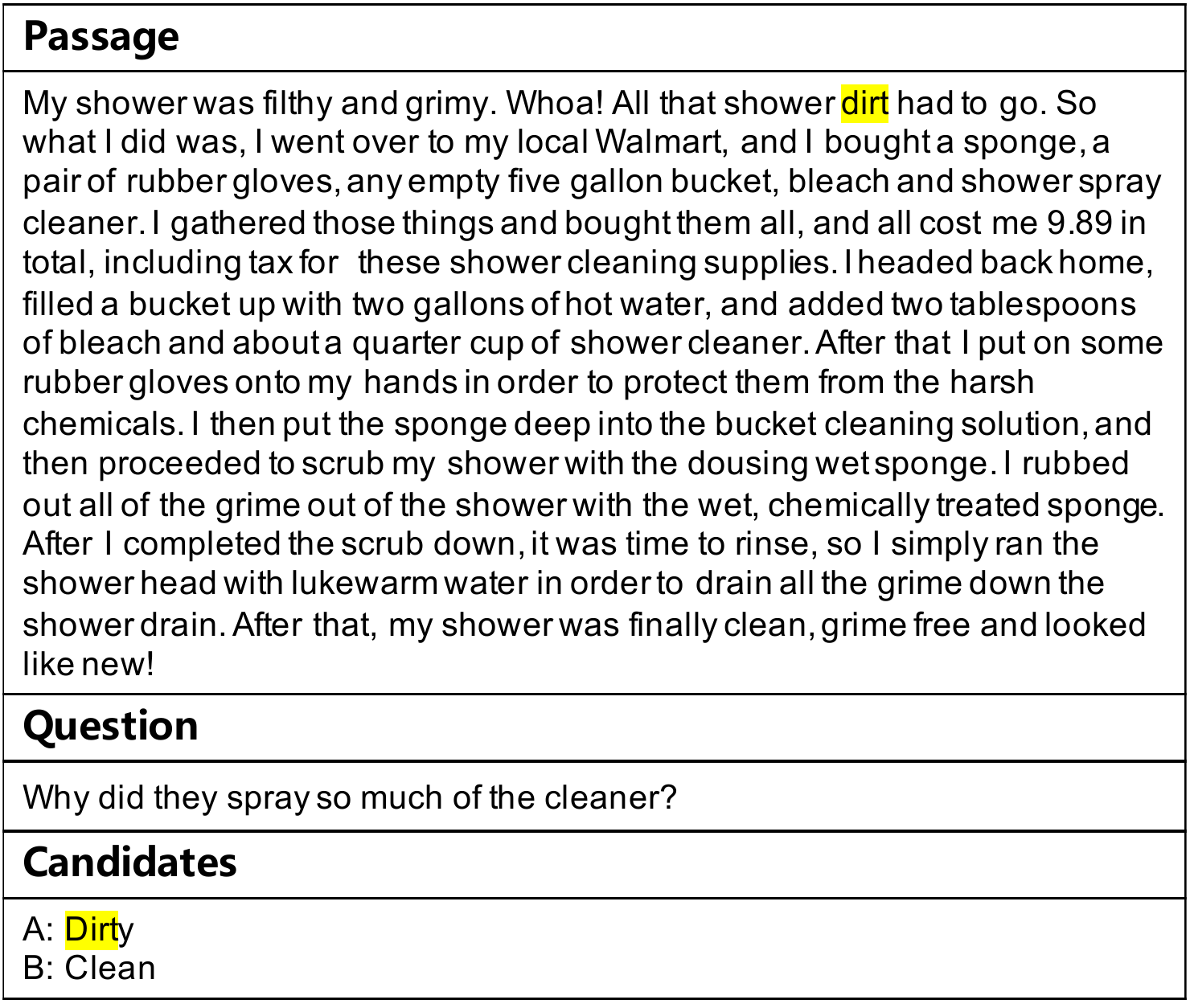}
\caption{\label{fig2} Example of the partial matching problem. }
\end{figure}

Another observation is that, different from SQuAD dataset \citep{rajpurkar-etal-2016}, the choices are not extracted from the text but written by the human. So there are many questions that the ordinary word matching failed to give correct answer. For example, the example shown in Figure \ref{fig2}, the choice `dirty' is not exactly match the word `dirt' in passage, which add difficulties in solving these problems. 
In our model, we add additional partial matching feature to indicate the underlying relations. The final results show that adding partial matching feature could give an improvement of 1\%$\sim$2\% in accuracy, indicating that the feature is helpful for the model to identify the words that have similar meanings.
Also, we have tried to use word stemming to restore the word to its stemming form, we observed a significant drop in the final performance, suggesting that a much powerful model is needed to further tackle these problems.

\section{Conclusion}
In this system description, we propose a novel neural network system called Hybrid Multi-Aspects (HMA) model for the SemEval-2018 Task 11.
In this model, we aim to produce multi-aspect output and combine them for final predictions.
We adopt a simple dot product to measure the similarity between the text, question and choices. 
We also enhanced the question representations by using self-attention mechanism. 
The final predictions are obtained by accumulating the probabilities from various aspects.
The final leaderboard provided by the task officials shows that the proposed HMA model got the first place among 24 participants with the accuracy of 84.13\%.
In the future, we would like to investigate how to effectively adopt the external knowledge into machine reading comprehension model and would like to focus on solving the questions that is less likely answered by statistical data.

\section*{Acknowledgments}
This work was supported by the National 863 Leading Technology Research Project via grant 2015AA015409. 

\bibliography{naaclhlt2018}
\bibliographystyle{acl_natbib}

\end{document}